\documentclass[11pt]{article}

\usepackage{ACL2023}

\usepackage{times}
\usepackage{latexsym}

\usepackage[T1]{fontenc}

\usepackage[utf8]{inputenc}

\usepackage{microtype}

\usepackage{inconsolata}

\usepackage{graphicx}
\usepackage{caption}
\usepackage{subcaption}
\usepackage{amsmath}
\usepackage{algorithm}
\usepackage{algpseudocode}
\usepackage{tcolorbox}
\usepackage{listings}
\usepackage{xcolor}
\usepackage{xspace} 
\usepackage{booktabs}
\usepackage{array}
\usepackage{listings}
\usepackage{amsfonts}
\usepackage{enumitem}
\usepackage{multirow}

\definecolor{codegreen}{rgb}{0,0.6,0}
\definecolor{codegray}{rgb}{0.5,0.5,0.5}
\definecolor{codepurple}{rgb}{0.58,0,0.82}
\definecolor{backcolour}{rgb}{0.95,0.95,0.92}

\newcommand{\name}{self-verified LLM\xspace}
\newcommand{\bchname}{SensorBench\xspace}
\newcommand{\equalcontrib}{\thanks{The research was conducted at UCLA and unrelated to the current affiliation.}}

\usepackage{listings}

\title{\bchname: Benchmarking LLMs in Coding-Based Sensor Processing}

\author{Pengrui Quan \\
  UCLA\\
  \texttt{prquan@g.ucla.edu} \\\And
  Xiaomin Ouyang\equalcontrib \\
  HKUST\\
  \texttt{xmouyang@cse.ust.hk} \\ \And
  Jeya Vikranth Jeyakumar\footnotemark[1]\\
  NVIDIA \\
  \texttt{vikranth94@g.ucla.edu} \\
  \AND
  Ziqi Wang\footnotemark[1] \\
  Qualcomm \\
  \texttt{wangzq312@g.ucla.edu} \\ \And
  Yang Xing \\
  UCLA \\
  \texttt{joker2003@g.ucla.edu}\And
  Mani Srivastava\thanks{The author holds concurrent appointments as Amazon Scholar and Professor at UCLA, but the work in this paper is not associated with Amazon.} \\
  UCLA \\
  \texttt{mbs@ucla.edu}
  }

\begin{document}
\maketitle
\begin{abstract}

Effective processing and interpretation of sensor data is crucial for cyber-physical and IoT systems, which traditionally require deep expertise in signal processing. However, recent research suggests that large language models (LLMs) show promise in processing sensory data, potentially serving as copilots for developing sensing systems. Despite this potential, systematic evaluation remains challenging due to fragmented studies with varied methodologies. Concerns also exist about LLMs' reliability for complex sensor processing tasks. To address these issues, we introduce SensorBench, a comprehensive benchmark to systematically evaluate LLMs' capabilities in sensor data processing. SensorBench incorporates diverse real-world sensor datasets across various tasks. Our results reveal that while LLMs perform well on simpler tasks, they struggle with more complex compositional tasks involving parameter selections compared to engineering experts. We also investigate the impact of different prompting strategies and fine-tuning approaches. Self-verification-based prompting achieves superior performance among all strategies, while supervised fine-tuning yields only limited improvements. This study provides a thorough benchmark and analysis to guide future developments in LLM-based sensor processing.\footnote{Code and data are available at \url{https://github.com/nesl/LLM_sensor_processing}.}

\end{abstract}

\section{Introduction}

\begin{figure}
    \centering
 \setlength{\abovecaptionskip}{0.cm}
    \setlength{\belowcaptionskip}{0.cm}
    \begin{subfigure}[b]{0.45\textwidth}
        \includegraphics[width=\textwidth]{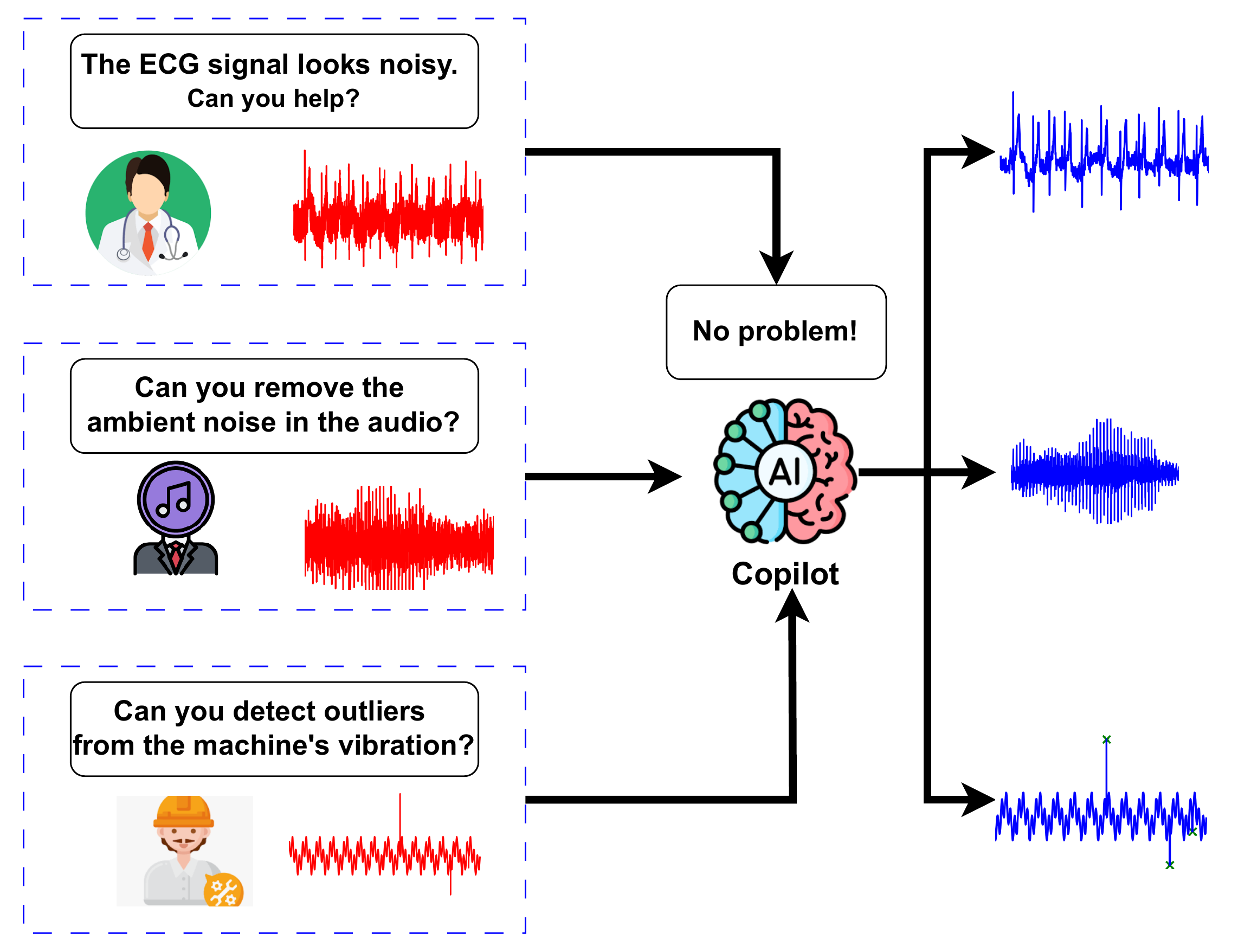}
    \end{subfigure}
    \caption{Sensor processing \emph{copilot}. We envision an intelligent assistant to support users, making advanced sensor data analysis accessible to broader audiences.}\label{fig:llmdsp}\vspace{-20pt}
\end{figure}

Recent advances in Large Language Models (LLMs) have sparked interest in their application to sensor signal processing and physical-world interactions. LLMs have shown promise in tasks such as analyzing sensory data for predictions related to health and human activity~\cite{ouyang2024llmsense, xu2023penetrative, kim2024health}. For example, \cite{ouyang2024llmsense} showed that the world knowledge acquired by LLMs during training makes them effective in diagnosing and monitoring the cognitive decline in Alzheimer’s patients from multimodal spatiotemporal sensor data from their homes. Likewise, \cite{xu2023penetrative, kim2024health, yoon2024my} showed that the knowledge of LLMs about the physical world enables them to interpret sensory data for human motion-related activities effectively.

Despite these early promising results, integrating LLMs into sensor processing systems faces significant challenges. The studies thus far are fragmented and use varied methodologies, datasets, and evaluation metrics. This prevents making definitive conclusions and reaching a cohesive understanding of LLM performance in this domain. Furthermore,  considerable doubts and skepticism persist about the reliability of LLMs for tasks like planning and reasoning \cite{stechly2024self, verma2024brittle}. Constructing an effective and efficient sensor processing pipeline is a type of planning and reasoning task, which requires an LLM to interpret the problem, generate a well-formed sequence of algorithmic steps, and execute them accurately. It is challenging to assess the full potential and limitations of LLM-based sensor processing without a structured approach.

Recognizing the need for a structured evaluation of LLMs' sensor processing abilities, this paper presents the first systematic attempt to bridge the gap. Specifically, we aim to answer the following two key research questions:

\noindent\textbf{(Q1) How well do LLMs perform sensor processing compared to human experts?} This analysis helps determine whether LLMs can serve as viable alternatives to human experts or if they fall short in key areas of the sensor processing workflow. 

\noindent\textbf{(Q2) Can prompt engineering and fine-tuning improve LLMs' performance in sensor processing?} We explore the effectiveness of advanced prompting strategies, such as Chain-of-Thought (CoT) \cite{wei2022chain}, ReAct \cite{yao2022react}, and self-verification \cite{shinn2024reflexion}. We aim to answer if LLMs can benefit from a reasoning approach inspired by the iterative problem-solving mindset of human experts. In addition, as fine-tuning is a common technique to improve models' performance, we investigate whether fine-tuning on a curated Digital Signal Processing (DSP) corpus will yield similar improvements.

To answer these questions, we first construct \bchname, a comprehensive benchmark comprising a diverse range of sensory types and tasks in realistic scenarios. Our evaluation provides answers to Q1, revealing that \textit{LLMs perform comparably to experts on simpler tasks but struggle with more complex compositional tasks and tasks that require parameter selection.}\looseness=-1

Furthermore, to enhance LLMs' capabilities as sensor processors, we explicitly employ existing state-of-the-art prompting strategies to guide LLMs to mimic the human "cognitive process." Out of the four prompting strategies we tested, the adapted self-verification approach demonstrated the best advantages, outperforming other baselines 48\% of the time. In contrast, fine-tuning using 11266 sensor processing challenges generated from 16 open-source tutorials and codebases shows little benefit. The evaluation results answer Q2, showing that \textit{LLMs benefit significantly from iterative sanity checks and reflection on the solutions, while knowledge injection through fine-tuning has limited improvements.}

Our main contributions are summarized below:
\begin{itemize}[leftmargin=2em]
    \item We propose a comprehensive benchmark for evaluating LLM performance in sensory data processing, providing a structured reference for future research. 
    \item We provide an in-depth comparison of LLM performance against human experts, highlighting the current state of LLM capabilities and identifying areas for improvement.
    \item We investigate various prompting strategies for sensor processing and demonstrate that self-verification \cite{shinn2024reflexion} is the best prompting method in 48\% of tasks. On the other hand, although fine-tuning is often used as a method for task-specific knowledge injection, we show that it yields only limited improvements. Our experiments further imply that the current LLMs intrinsically perform knowledge retrieval rather than planning and reasoning without external guidance.\looseness=-1
\end{itemize}
\section{Related Works}

\noindent{\bf LLM for Sensor Data Analysis.} \cite{xu2023penetrative} used LLMs to process ECG signals and satellite WiFi SSID signals, showing their ability to fuse and reason with sensor data. Similarly, \cite{ouyang2024llmsense} applied LLMs to high-level reasoning tasks using household sensor data, such as air quality and occupancy sensors, to diagnose Alzheimer's disease. Other studies \cite{ji2024hargpt, yoon2024my} explored human activity recognition using IMU sensors. These works highlight LLMs' potential to generalize across sensor tasks without task-specific training. However, there is a lack of systematic investigations, as research methods and evaluation metrics remain inconsistent across different studies.\looseness=-1


\noindent{\bf Tool-augmented LLMs.} Considerable research has focused on building LLMs to tackle general coding challenges \cite{shojaee2023execution, dubey2024llama, nijkamp2022codegen, luo2023wizardcoder}. However, the models' abilities to solve real-world sensor processing tasks remain questionable and have not been systematically studied. Moreover, although efforts have gone into building LLM-based agents using existing tool interfaces \cite{yue2024mmmu, ge2024openagi, qin2023toolllm, patil2023gorilla, zhuang2023toolqa}, most of these works have centered on tasks like video questioning, image generation, and object detection -- areas supported by robust APIs that do not require extensive domain-specific knowledge. 


\noindent{\bf Benchmarking LLMs.} There are two genres of benchmarks related to our work. Firstly, \cite{tan2023towards, jin2023time, dhingra2022time, chen2021dataset} evaluated the spatio-temporal reasoning capability of LLMs. These studies only focused on the reasoning capability in a natural language context, such as temporal concepts, causality, or sequential orders. However, processing sensor signals is more complex, which requires working with physical-world signals and professional processing pipelines like filtering. Secondly, \cite{chen2021evaluating, du2023classeval} evaluated LLM's coding capability to manipulate tools. However, few of them directly dealt with processing real-world sensor signals, preventing making conclusive claims on the LLMs' capability in sensor processing abilities.

\section{\bchname}\label{sec:benchmark}
\begin{table*}[ht]
\centering
\resizebox{0.95\textwidth}{!}{%
\begin{tabular}{@{}lllllllll@{}}
\toprule
Category          & Task                                                  & Dataset         & Modality               & Metric & Single/Comp. & Non-Para./Para. & Difficulty & Percentage (\%) \\ 

\midrule
Preprocessing     & \begin{tabular}[c]{@{}l@{}}

 Resampling\\ Delay detection\end{tabular} & \begin{tabular}[c]{@{}l@{}} Synthetic \\Gait \cite{luo2020database} \end{tabular} & \begin{tabular}[c]{@{}l@{}} Synthetic \\Pressure \end{tabular} & MSE & Single & Non-Para. & 2 & 8   \\
 
\midrule
Signal reconstruction &
  \begin{tabular}[c]{@{}l@{}}
    Imputation \\ 
    Extrapolation
  \end{tabular} &
  \begin{tabular}[c]{@{}l@{}}
  ECG\cite{goldberger2000physiobank}\\
  PPG\cite{zhang2014troika}
  \end{tabular} & ECG \& PPG & MSE & Single & Para. & 3 & 16
    \\
\midrule
Spectrum analysis &
  \begin{tabular}[c]{@{}l@{}}
    
    Period detection 
  \end{tabular} &
  \begin{tabular}[c]{@{}l@{}}
    Gait \cite{luo2020database}
  \end{tabular} & Pressure &
  MSE & Single & Para. & 3 & 4 \\

\midrule 
Outlier detection & \begin{tabular}[c]{@{}l@{}}Additive outlier removal\\ \end{tabular}                   & 
    \begin{tabular}[c]{@{}l@{}} Yahoo  \cite{laptev2015yahoo}\\
    Additive noise \\
    Shifted distribution\\
    Shifted mean
    \end{tabular} & \begin{tabular}[c]{@{}l@{}} Network traffic \\ Synthetic
    \end{tabular}
    & F1 score & Comp. & Para. & 4 & 16 \\
                            
\midrule
Spectral filtering &
  \begin{tabular}[c]{@{}l@{}}
    Low stop \\
    Band stop \\
    Band pass \\
    High stop
  \end{tabular} &
  \begin{tabular}[c]{@{}l@{}}ECG-baseline drift \\ 
    Spoke digits - noise \cite{jackson2018jakobovski} \\
    ECG-Powerline\\
    ECG-Gaussian
    \end{tabular} & Audio \& ECG &
  \begin{tabular}[c]{@{}l@{}} MSE \\ SDR \\ MSE \\ MSE \end{tabular} & \begin{tabular}[c]{@{}l@{}} Single \\ comp. \\ Single \\ Single \end{tabular} & Para. & \begin{tabular}[c]{@{}l@{}} 3 \\ 4 \\ 3 \\ 3 \end{tabular} & 32 \\

\midrule
Adaptive filtering &
  \begin{tabular}[c]{@{}l@{}}
    Echo cancellation
  \end{tabular} &
  \begin{tabular}[c]{@{}l@{}} Spoke Digits \cite{jackson2018jakobovski}\end{tabular} & Audio &
  \begin{tabular}[c]{@{}l@{}}  SDR \end{tabular} & Comp. & Para. & 4 & 4 \\

\midrule

Feature extraction &
  \begin{tabular}[c]{@{}l@{}}
    Heart rate calculation \\
  \end{tabular} &
  \begin{tabular}[c]{@{}l@{}}
    ECG\cite{goldberger2000physiobank} \\
  \end{tabular} & ECG &
  \begin{tabular}[c]{@{}l@{}}
    MSE
  \end{tabular} & Comp. & Para. & 4 & 4 \\
  
\midrule
Change point detection &

  Change point detection &
  \begin{tabular}[c]{@{}l@{}}
    Changing variance \\
    Changing means\\
    Changing frequency
  \end{tabular} & Synthetic  & F1 score & Comp. & Para. & 4 & 16
  \\
\bottomrule
\end{tabular}
}
\caption{Overview of sensor processing tasks and datasets.}\label{table:benchmark_org}
\vspace{-15pt}
\end{table*}
Given the heterogeneity of recent studies on evaluating LLMs for sensor signal processing \cite{kim2024health, yoon2024my, xu2023penetrative}, we propose \bchname for a more structured study.\looseness=-1

\subsection{Benchmark Overview}

\bchname focuses on processing temporal sensor signals, such as audio, ECG, PPG, motion, and pressure signals. Since sensor processing typically involves digital signal processing (DSP) methods, we start by selecting DSP tasks common in engineering and industrial settings. We select examples from MATLAB tutorials \cite{matlab_dsp_examples, matlab_signal_examples, matlab_signal_smoothing_examples}, leveraging their practical and instructional approach, and subsequently refer to established DSP textbooks \cite{oppenheim1999discrete, roberts1987digital} to create our benchmark. This method ensures that we capture real-world applications and validate them against academic references.\looseness=-1 

The benchmark is organized by categorizing the difficulty of each task based on two primary criteria: single (Diff. 1) or compositional (Diff. 2) and non-parameterized (Diff. 1) or parameterized (Diff. 2), as summarized in Table \ref{table:benchmark_org}.

\begin{figure}
    \centering
 \setlength{\abovecaptionskip}{-0.5cm}
    \setlength{\belowcaptionskip}{0.cm}
        \includegraphics[width=\linewidth]{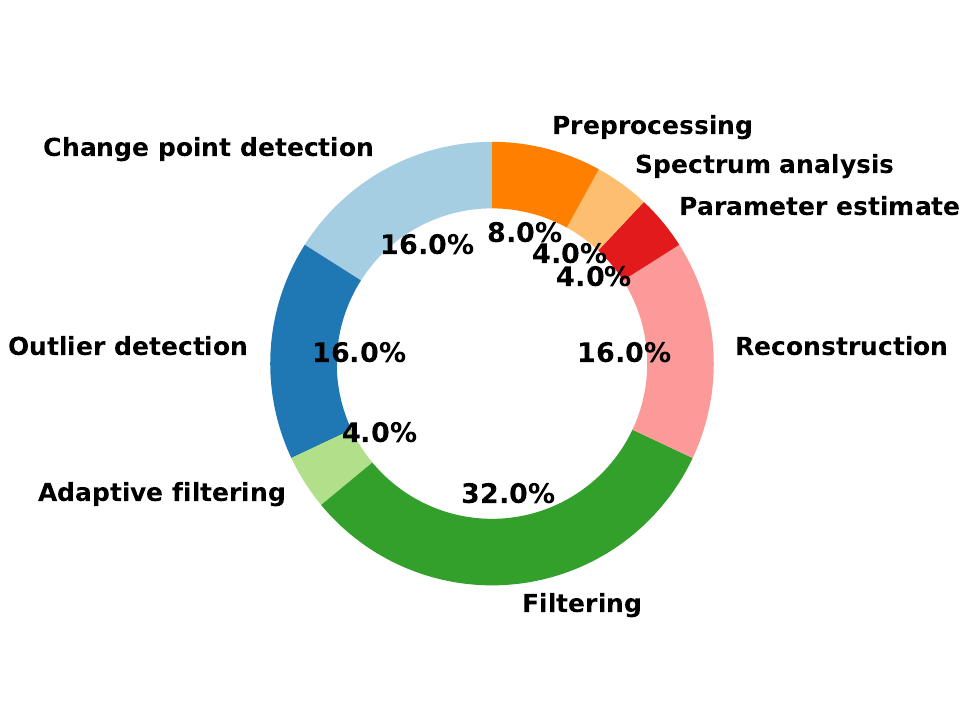 }
    \caption{Benchmark organization by categories.} 
    \label{fig:dataset_org}
    \vspace{-20pt}
\end{figure}

\noindent{\textbf{Single/Compositional.}} Firstly, single tasks involve straightforward, isolated API calls. In contrast, compositional tasks require coordinating several interconnected processes or interpreting signals from multiple tools to achieve a unified goal.

\noindent{\textbf{Parameterized/Non-parameterized.}} Parameterized tasks involve setting specific critical parameters that determine the processing quality, such as setting a stop-band frequency for spectral filtering tasks. Non-parameterized tasks, in contrast, do not involve parameter selection.

We calculated the overall difficulty for each task by summing the two factors. Overall, Table \ref{table:benchmark_org} summarizes \bchname, with the share of each category in the benchmark composition.

\subsection{Tasks and Datasets Generation}

 We introduce tasks and dataset generation procedures in \bchname as follows:


\textbf{Preprocessing. } We investigate two preprocessing tasks: (1) \emph{Resampling} involves changing a signal's sampling rate. Here, we designed synthetic data with compositions of sine waves and performed sub-Nyquist downsampling. (2) \emph{Delay detection} refers to the process of detecting the delay between two signals due to asynchronous data collection or time delay caused by wave propagation. We manually inject delays into gait signals \cite{luo2020database} to simulate the task.

\textbf{Signal reconstruction.} We focus on extrapolation and imputation. We mask data to generate missing data from PPG \cite{zhang2014troika} and ECG \cite{goldberger2000physiobank} for \emph{imputation} and truncate the original signal for \emph{extrapolation}. We randomly mask 25 samples from 3-second ECG and PPG signals accordingly.

\textbf{Spectrum analysis.} We mainly employ \emph{periodicity detection} to evaluate the ability of frequency analysis. We use the gait signals \cite{luo2020database} and use the example code in MATLAB \cite{mathworks_periodicity} to generate ground truth periodicity.

\textbf{Outlier detection.} We take the existing Yahoo network traffic dataset \cite{laptev2015yahoo} and generate synthetic datasets combining additive trends, periodicity, and missing values. 

{\bf Filter design.} We evaluate the filter design ability using ECG and audio.

\noindent{\textit{Spectral filter - ECG.}} ECG signals could be corrupted by various types of noise while being sampled at different rates. We use the original ECG data \cite{goldberger2000physiobank} as ground truth and inject the following noise to generate the corrupted versions.\looseness=-1

(1) \emph{Baseline Drift / Detrend (Low Stop)}: The baseline drift is caused by the respiration or movement of a subject. (2) \emph{Powerline Noise (Bandstop)}: The interference from mains electricity (50 or 60 Hz) can corrupt physiological signals. (3) \emph{Gaussian noise (High Stop)}: Gaussian noise has constant power spectral density across all frequencies. 

\noindent{\textit{Spectral filter - Speech.}} We use the spoken digit dataset \cite{jackson2018jakobovski} as ground truth. We injected noise with the following combinations of characteristics:

(1) \emph{Single peak v.s. Multi-peak.} Noise with a single peak is focused around a specific frequency, whereas noise with multiple peaks is more dispersed and complex.
(2) \emph{Stationary vs. Non-Stationary Noise.} Stationary noise is predictable and easier to filter, while non-stationary noise has properties that change over time. 

\textbf{Adaptive Filtering} continuously updates their coefficients to minimize a defined error criterion. We use the task of \emph{Echo Cancellation} to evaluate models' ability to perform adaptive filtering. We use the spoken digit dataset \cite{jackson2018jakobovski} and generate the corrupted versions with echo.


\textbf{Parameter estimation.} We use \emph{Heart rate calculation} in ECG signals as an example, where it mainly refers to identifying the corresponding R-peak and calculated heart rate. We use the R-peak labels in \cite{goldberger2000physiobank} as ground truth.

{\textbf{Change point detection} involves the identification of abrupt changes or discontinuities in the statistical properties. We generate data with various property changes, including mean, variance, and frequency change, and record the change points as ground truth.}\looseness=-1

In summary, the \bchname comprises 11 distinct tasks across 9 different data modalities, yielding 25 unique modality-task pairs. For each pair, we generate 10 signals, resulting in a total of 250 sensor processing challenges (a subset of 75 challenges is utilized in this work to compare with human experts). Figure \ref{fig:dataset_org} illustrates the benchmark’s organization by category, and Table \ref{table:benchmark_org} provides a summary of these tasks.

\begin{figure*}
    \centering
 \setlength{\abovecaptionskip}{0.cm}
    \setlength{\belowcaptionskip}{0.cm}
    \begin{subfigure}[b]{0.9\textwidth}
        \includegraphics[width=\textwidth]{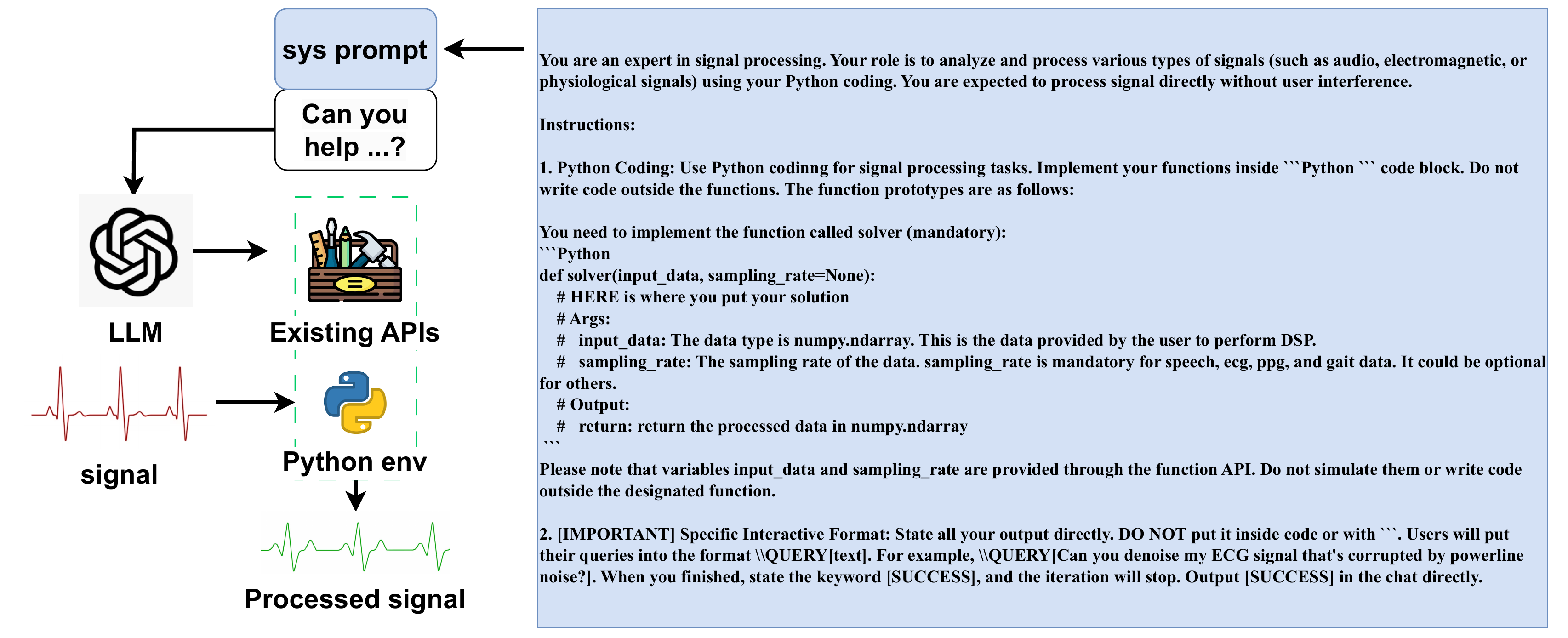}
    \end{subfigure}

    \caption{Coding with APIs. System prompts and user queries are fed to LLMs as instructions. The models can access and process signals through \textit{Python}, with the help of defined APIs.} 
    \label{fig:setting_LLMDSP}
\vspace{-10pt}
\end{figure*}

\section{Evaluating LLMs on Sensorbench}\label{Sec:evaluate_llm}

We benchmark four leading LLMs - GPT-4o \cite{gpt_4o}, GPT-4 \cite{achiam2023gpt}, GPT-3.5-turbo \cite{gpt_3_5_turbo}, and Llama-3.1-70b \cite{dubey2024llama} - on SensorBench to assess their capabilities in sensor data processing tasks. This evaluation aims to answer our first research question (Q1) about how well LLMs perform compared to human experts.

\subsection{Experiment Setup}

\subsubsection{Evaluation Metrics}
We employ the following metrics (also shown in Table. \ref{table:benchmark_org}) to comprehensively evaluate performance across different tasks: 

\noindent{\bf Signal-to-Distortion Ratio (SDR).} Used for tasks like signal reconstruction and processing. A higher SDR indicates better performance, as it reflects how accurately the signal was processed or reconstructed.\looseness=-1

\noindent{\bf F1 Score.} Utilized for detection tasks to balance precision and recall. This metric is crucial for tasks like change point detection and outlier detection, where both false positives and false negatives are important considerations.

\noindent{\bf Mean Squared Error (MSE).} Applied to measure the closeness between processed signals and ground truth. A lower MSE indicates better performance, making it suitable for tasks where exact signal matching is critical.

\noindent{\bf Win rate.} Calculated as the percentage of tasks where an LLM achieves the best performance among all tested models. This metric helps identify which models consistently outperform others across various tasks.

\begin{equation} 
\small
    \text{win rate} = \frac{\sum_{i=1}^{N} \mathbb{I}(\text{output is the best for task } i)}{N} \times 100\%
\end{equation}

\noindent{\bf Failure rate.} Measures the percentage of invalid outputs produced by an LLM. This metric is essential for assessing the reliability and robustness of models in real-world applications.

\begin{equation}
\small
    \text{failure rate} = \frac{\sum_{i=1}^{N} \mathbb{I}(\text{output is invalid for task } i)}{N} \times 100\%
\end{equation}

\subsubsection{Settings of Leveraging LLMs}

\begin{table}
\centering
\resizebox{0.48\textwidth}{!}{%

\begin{tabular}{lccc}
\toprule
 Models & coding w/ APIs (A) & coding w/o APIs (NA) & Text-based (T)\\
\midrule
GPT-4o & \textbf{60.0} & 28.0 & 12.0 \\
GPT-4 & \textbf{56.0} & 24.0 & 20.0 \\
GPT-3.5-turbo & \textbf{56.0} & 20.0 & 24.0 \\
Llama-3-70b & \textbf{68.0} & 20.0 & 12.0 \\
\bottomrule
\end{tabular}
}
\caption{Average win rate. We compute the percentage of a setting winning for the same model and take the average over all tasks.}
\vspace{-15pt}
\label{table:setings}
\end{table}

To explore using LLMs for sensor processing, we designed three distinct interaction settings that mimic expert engineers' ways of processing sensors. (1) Coding with APIs: LLMs leverage pre-defined libraries and functions for DSP tasks. This is similar to how engineers typically approach these problems using established tools.
(2) Coding without APIs: LLMs generate custom code solutions independently. 
(3) Text-based interaction: LLMs tackle DSP challenges through conversation. 
We evaluated each setting and observed that LLMs perform the best under the setting of coding with APIs as shown in Table \ref{table:setings} (more details are in Appendix \ref{appendix:settings}), and hence the remaining part of our paper focuses on the strategy. Figure \ref{fig:setting_LLMDSP} shows the system diagram of the ``coding with APIs'' setting. With user queries and targeted signals provided to the LLMs, we instruct the models to leverage pre-defined libraries and functions for DSP challenges (the defined list of available APIs includes \emph{numpy}, \emph{scipy}, \emph{pandas}, \emph{pmdarima}, \emph{statsmodels}, and \emph{ruptures} \cite{harris2020array, virtanen2020scipy, mckinney2010pandas, pmdarima2020, statsmodels2010, ruptures2018}). 

\subsubsection{Evaluation Protocol}


To ensure robust and reliable benchmarking, we implemented a rigorous evaluation protocol:

For each task-dataset combination, we selected 3 distinct signals and repeated each experiment 3 times, resulting in 9 examples per combination.
We calculated a trimmed mean by dropping the highest and lowest scores from the 9 examples and averaging the remaining 7. This approach mitigates the impact of outliers, which is crucial when dealing with unbounded metrics like MSE and SDR.
This protocol ensures that our results are representative and not skewed by random variations or extreme values.\looseness=-1

\subsection{Comparison with Human Experts}

\begin{figure*}[htbp]
    \centering
 \setlength{\abovecaptionskip}{0.cm}
    \setlength{\belowcaptionskip}{0.cm}
    \begin{subfigure}[b]{0.24\textwidth}
        \includegraphics[width=\textwidth]{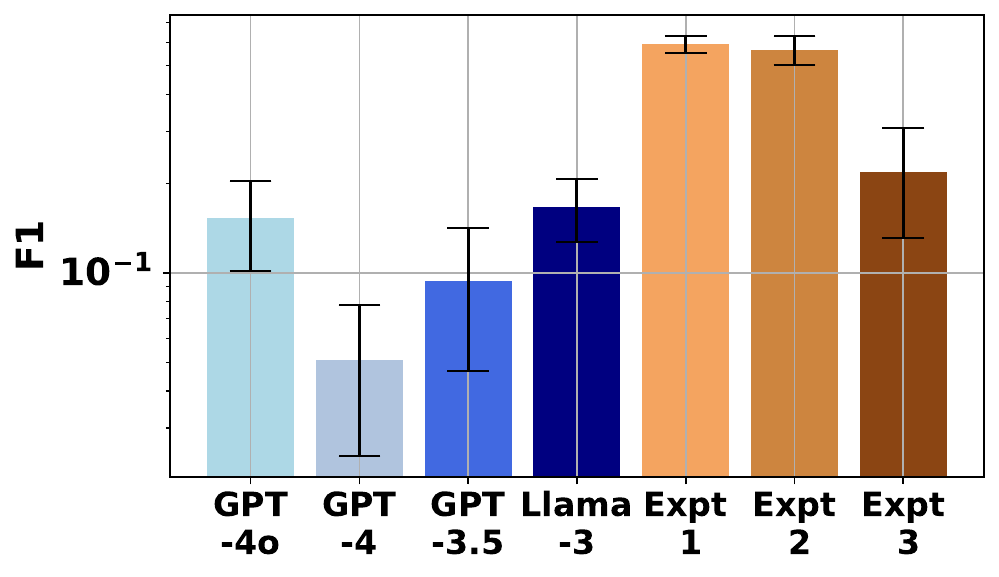}
        \caption{Change point detection}
        \label{fig:change_point_detection}
    \end{subfigure}
    \begin{subfigure}[b]{0.24\textwidth}
        \includegraphics[width=\textwidth]{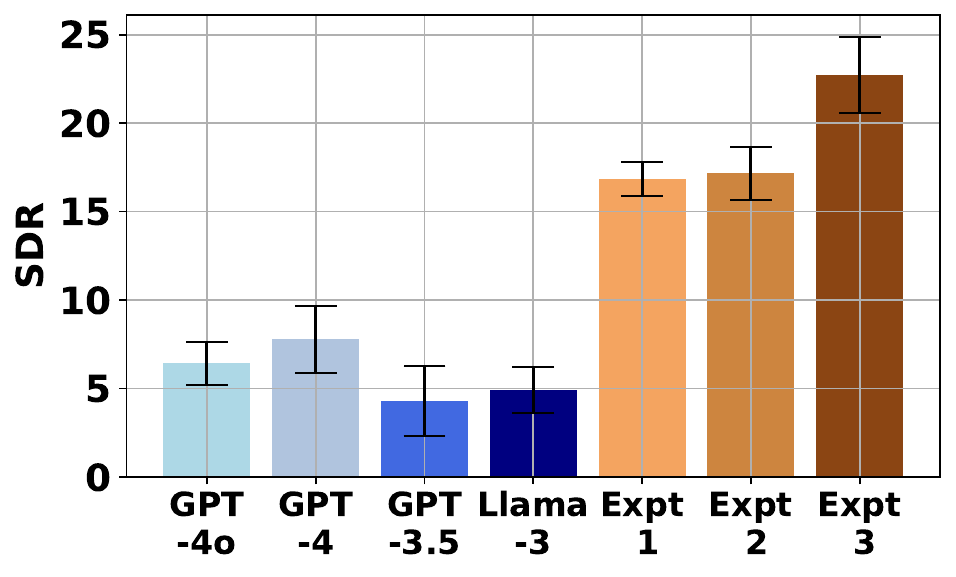}
        \caption{Spectral filtering-speech}
        \label{fig:filtering_speech}
    \end{subfigure}
    \begin{subfigure}[b]{0.24\textwidth}
        \includegraphics[width=\textwidth]{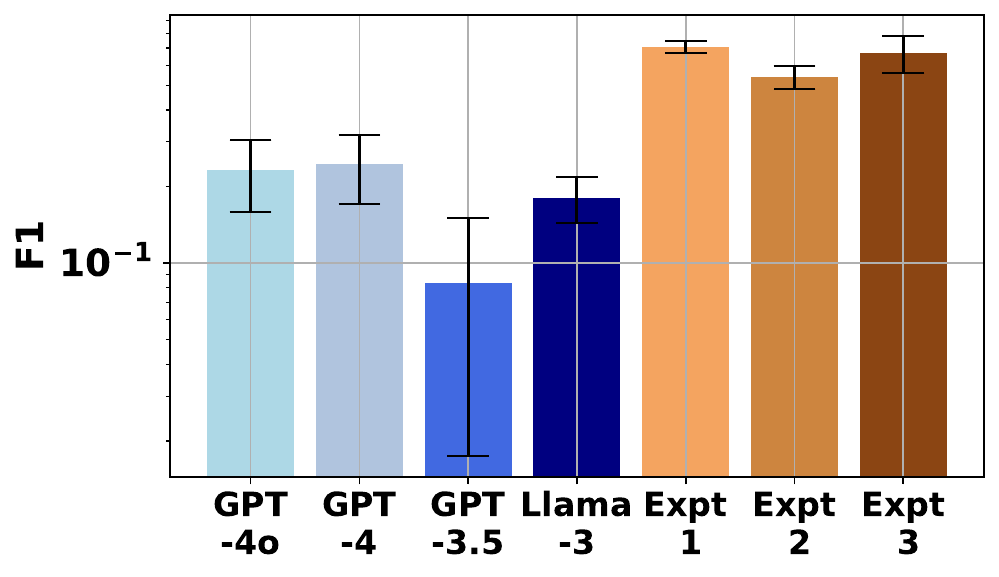}
        \caption{Outlier detection}
        \label{fig:outlier_detection}
    \end{subfigure}
    \begin{subfigure}[b]{0.24\textwidth}
        \includegraphics[width=\textwidth]{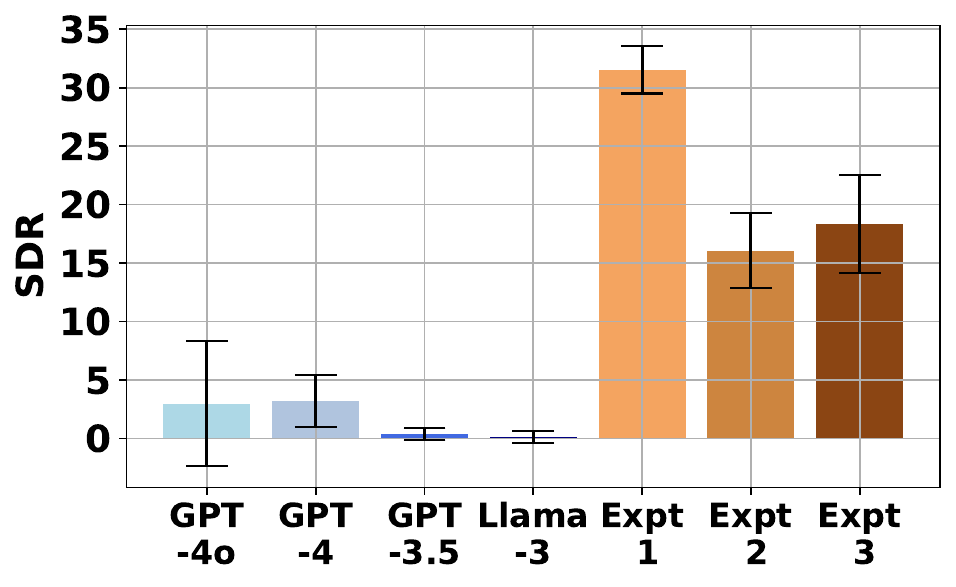}
        \caption{Echo cancellation}
        \label{fig:echo_cancellation}
    \end{subfigure}

    \begin{subfigure}[b]{0.24\textwidth}
        \includegraphics[width=\textwidth]{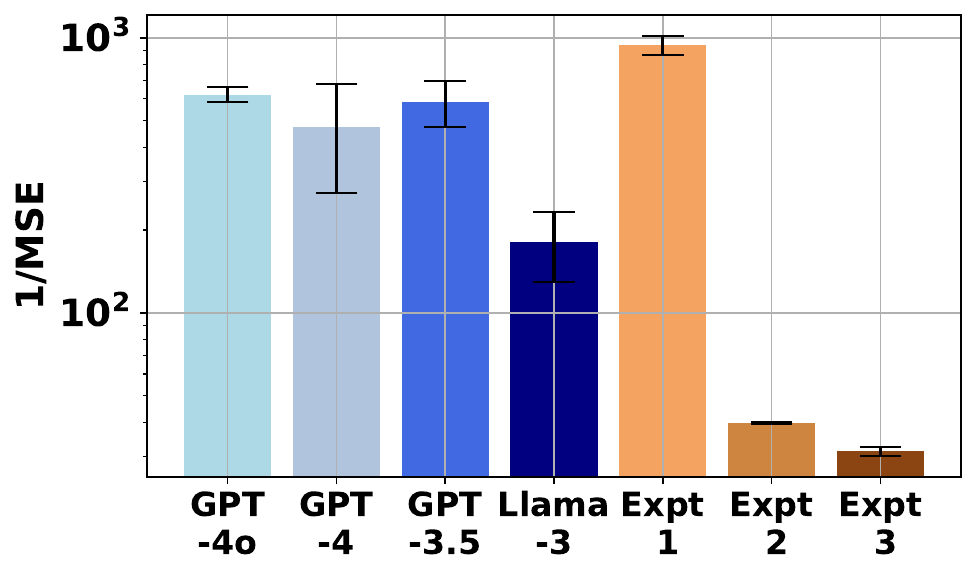}
        \caption{Spectral filtering-ECG}
        \label{fig:Spectral filtering-ECG}
    \end{subfigure}
    \begin{subfigure}[b]{0.24\textwidth}
        \includegraphics[width=\textwidth]{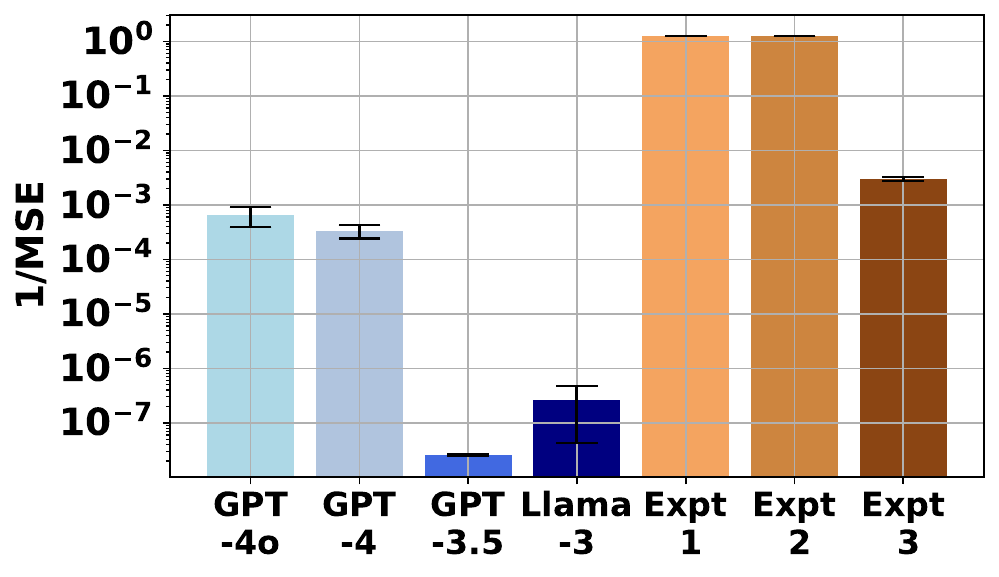}
        \caption{Heart rate calculation}
        \label{fig:heart_rate}
    \end{subfigure}
    \begin{subfigure}[b]{0.24\textwidth}
        \includegraphics[width=\textwidth]{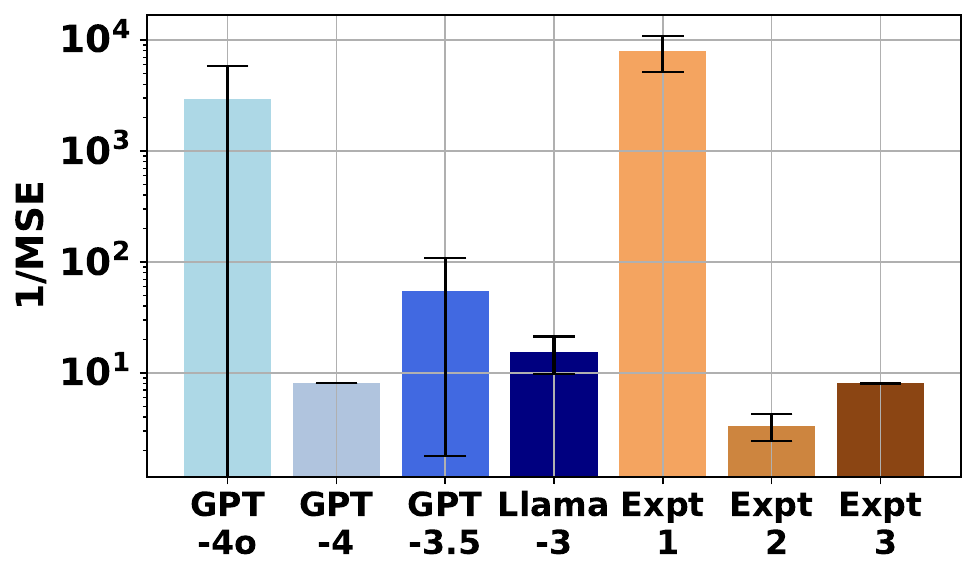}
        \caption{Resampling}
        \label{fig:resampling}
    \end{subfigure}
    \begin{subfigure}[b]{0.24\textwidth}
        \includegraphics[width=\textwidth]{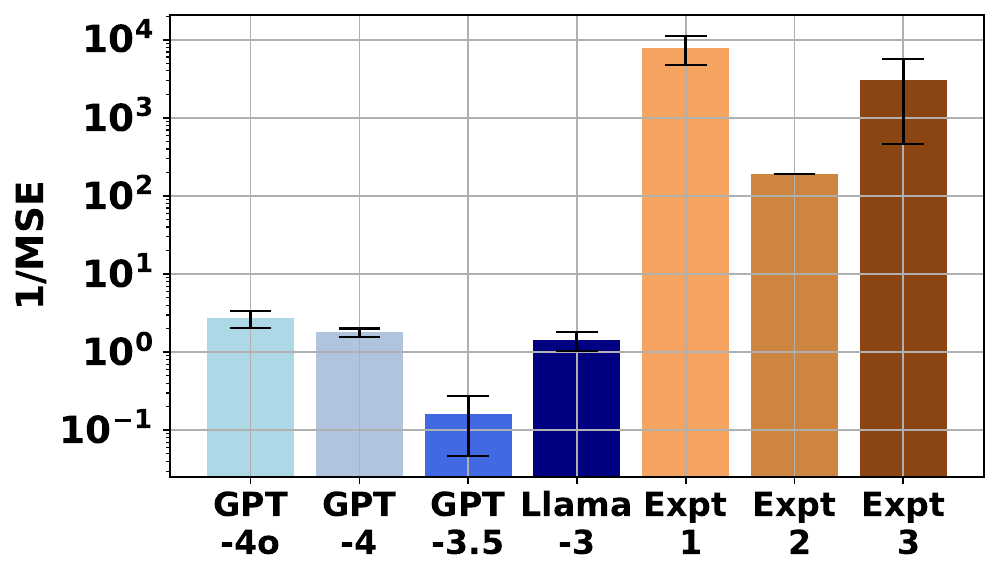}
        \caption{Period detection}
        \label{fig:Period detection}
    \end{subfigure}

        \begin{subfigure}[b]{0.24\textwidth}
        \includegraphics[width=\textwidth]{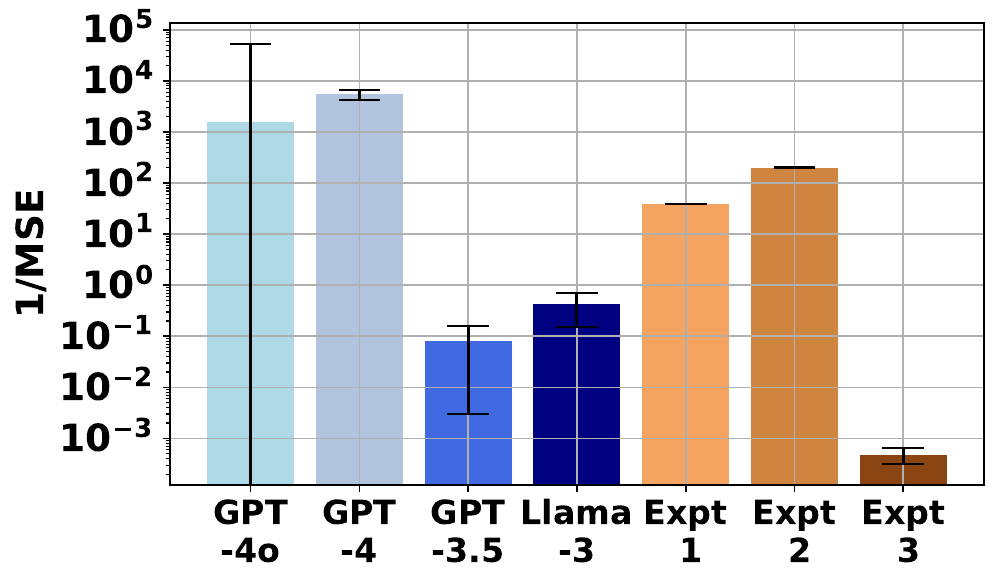}
        \caption{Delay detection}
        \label{fig:Delay detection}
    \end{subfigure}
    \begin{subfigure}[b]{0.24\textwidth}
        \includegraphics[width=\textwidth]{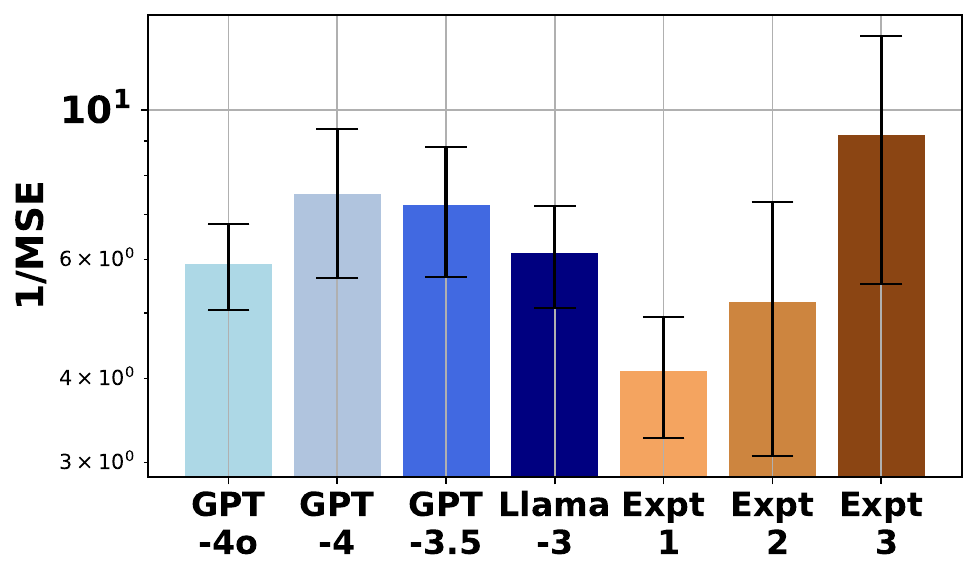}
        \caption{Imputation}
        \label{fig:imputation}
    \end{subfigure}
    \begin{subfigure}[b]{0.24\textwidth}
        \includegraphics[width=\textwidth]{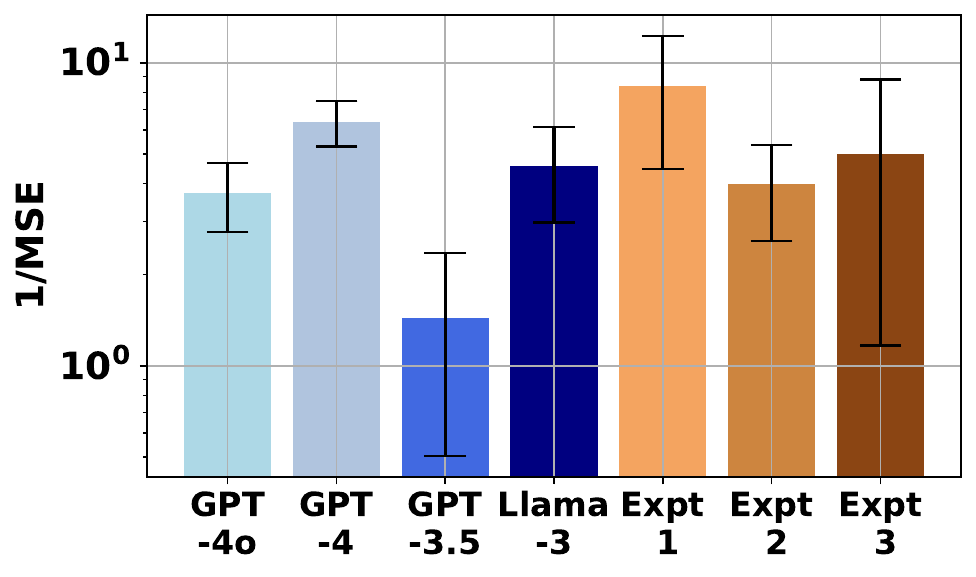}
        \caption{Extrapolation}
        \label{fig:extrapolation}
    \end{subfigure}

    \caption{LLMs v.s. Experts. For the task of using MSE as a metric, we use $1/\text{MSE}$ to ease the comparison. The higher the number in the y-axis, the better the performance. On task (a) - (d), (f), and (h), human experts are substantially better than LLMs.}
    \label{fig:performance_comparison_expert}
    \vspace{-15pt}
\end{figure*}

\begin{table}
\centering
\resizebox{0.5\textwidth}{!}{%
\large
\begin{tabular}{>{\raggedright\arraybackslash}p{0.15\textwidth} >{\raggedright\arraybackslash}p{0.27\textwidth} >{\raggedright\arraybackslash}p{0.27\textwidth} >
{\raggedright\arraybackslash}p{0.27\textwidth}}
\toprule
 Demo. & Expert 1 & Expert 2 & Expert 3 \\
\midrule
Exp. & 8 yrs & 4 yrs & 3 yrs \\
\midrule
Background  & PhD in electrical and computer engineering. Incoming research scientist in a tech company & Junior year PhD student in electrical and computer engineering & Junior year undergrad in electrical and computer engineering \\
\midrule
Relevant courses taken & 
Probability and Statistics,
Signals and Communication Systems,
Digital Signal Processing,
Digital Speech Processing
 & Signals and Systems (Discrete/Continuous), Probability and Statistics, Modern Communication System, Speech Processing & Signals and Systems (Discrete/Continuous), Probability and Statistics, Intro to communication\\
\midrule
Field & Substantial research publications in wireless signals processing & Multimodal sensor fusion for real-time scene analysis. Publications in sensor processing & Time series data analysis\\
\midrule
Time & 11.8h & 10.5h & 7h\\
\midrule
Win rate  & 64\% & 56\% & 52\%\\
\bottomrule
\end{tabular}}
\caption{Background of the human experts. The win rate indicates the percentage of tasks in which the expert defeats all LLMs.}
\label{table:expert domains}\vspace{-20pt}
\end{table}

To establish a meaningful baseline, we compared LLM performance against three human experts in sensor processing. We gave LLMs and experts access to \emph{Python} and the APIs access (coding with APIs). All experts are well-educated, have taken relevant courses, and two of them had relevant publications. Table \ref{table:expert domains} shows the experts' relevant background\footnote{The study is IRB-approved (IRB protocol number omitted for anonymity).}.\looseness=-1

Table. \ref{table:expert domains} shows the win rates for both experts, defined as the percentage of tasks in which the expert outperforms all LLMs. Expert 1 surpasses all LLMs in 64\% of the tasks, Expert 2 does so in 56\% of the tasks, and Expert 3 wins 52\% of the tasks. Fig. \ref{fig:performance_comparison_expert} shows the taskwise performance for LLMs and the domain experts (we show $1/$MSE to ease the comparison).

\emph{Experts are better: Change point detection, Spectral filtering-speech, Outlier detection, Echo cancellation, Heart rate calculation, and Period detection.} Figure \ref{fig:performance_comparison_expert} demonstrates that all experts outperform the LLMs in these tasks significantly. These tasks usually require multi-step planning, iterative problem-solving, and parameter selection, which are deemed medium or high difficulty, as shown in Table. \ref{table:benchmark_org}.


\emph{Experts tie: Spectral filtering-ECG, Resampling, and Delay detection, Imputation, and Extrapolation.} As shown by figure \ref{fig:performance_comparison_expert}, LLMs outperform or achieve comparable performance to the human experts in these tasks. It is worth noticing that these tasks can be done by calling APIs with existing world knowledge obtained through the training corpus. For example, in Spectral filtering - ECG, the model can rely on \textit{scipy} to design a bandpass filter from 0.5 Hz to 45 Hz to remove almost all possible noise in ECG.



Besides, experts take a significant amount of time to solve these tasks - all experts spend 7 to 11 hours. In contrast, the average processing time for Llama-3-70B in the coding with APIs setting is 78 seconds per challenge \footnote{The time is measured on an NVIDIA H100 GPU using Unsloth libraries \cite{unsloth}.}.


Based on the results, we answer Q1 - while the current LLMs can achieve comparable performance on simple tasks (difficulties $\leq3$) in our benchmark, there are significant gaps between harder tasks (difficulties $>4$ ) that require iterative problem-solving and parameter tuning. 

The evaluation shed light on the challenges of \emph{sensor processing copilots} - LLMs are prone to errors, commonly referred to as hallucinations and confabulations, leading to task failures. Our qualitative analysis of various failure cases revealed two main causes: \textit{implementation errors} and \textit{inappropriate parameter selection}. Implementation errors, such as incorrect function template calls, can be effectively mitigated by providing execution error messages to LLMs. However, parameter selection poses a greater challenge. For instance, in an audio denoising task in Figure \ref{fig_filer_example}, the LLM might incorrectly assume the stop-band frequency, which requires careful audio spectrum analysis and feedback analysis. This limitation can be attributed to the limitation of the LLM's reasoning ability, as the necessary information about the stop band cannot be retrieved from training data and must be derived directly from the signal. The observations further motivate us to study the effects of prompting engineering and fine-tuning in Section \ref{Sec:evaluate_llmdsp}.

\section{Can Prompt Engineering and Fine-tuning Help?}\label{Sec:evaluate_llmdsp}

\begin{table*}[ht]
\centering
\resizebox{0.95\textwidth}{!}{%
\begin{tabular}{ll|l|cccc|cc|cc}
\toprule
 \multicolumn{3}{c|}{Models} & \multicolumn{4}{c|}{GPT-4o} & \multicolumn{2}{c|}{Llama-70B} & \multicolumn{2}{c}{Llama-3B} \\
\cmidrule(lr){1-3}\cmidrule(lr){4-7}\cmidrule(lr){8-9}\cmidrule(lr){10-11}
Task & Data characteristic & Metric & Base & CoT & ReAct & Self-verification & Base & Finetuned & Base & Finetuned \\
\midrule
$\uparrow$ Change point detect & Synth.-frequenncy & F1 & 0.1193 & 0.0915 & \textbf{0.1671} & \underline{0.1488} & \underline{0.0963} & \textbf{0.1029} & \underline{0.0408} & \textbf{0.0900} \\
$\uparrow$ Change point detect & Synth.-variance & F1 & \underline{0.0726} & 0.0115 & 0.0346 & \textbf{0.0809} & \textbf{0.1470} & \underline{0.0541} & \underline{0.0835} & \textbf{0.1260} \\
$\uparrow$ Change point detect & Synth.-mean & F1 & 0.3088 & \textbf{0.3912} & 0.2987 & \underline{0.3881} & \textbf{0.4246} & \underline{0.1932} & \underline{0.0495} & \textbf{0.1263} \\
$\uparrow$ Change point detect & Synth.-mean-noise & F1 & 0.1111 & 0.0952 & \underline{0.1481} & \textbf{0.4325} & \underline{0.0000} & \textbf{0.0472} & \underline{0.0667} & \textbf{0.1800} \\
$\uparrow$ Outlier detection & Synth.-trend-noise & F1 & 0.0207 & \underline{0.0581} & 0.0400 & \textbf{0.4588} & \underline{0.3333} & \textbf{0.4523} & nan & \textbf{0.2659} \\
$\uparrow$ Outlier detection & Synth.-noise-vary distr. & F1 & 0.0508 & 0.0571 & \underline{0.0989} & \textbf{0.2583} & \underline{0.0140} & \textbf{0.2000} & \underline{0.0144} & \textbf{0.0800} \\
$\uparrow$ Outlier detection & Synth.-noise & F1 & \underline{0.2738} & 0.2417 & \underline{0.2738} & \textbf{0.4729} & \textbf{0.1842} & \underline{0.1646} & nan & \textbf{0.0557} \\
$\uparrow$ Outlier detection & Network traffic & F1 & 0.5817 & \textbf{0.6845} & \underline{0.6171} & 0.3170 & \underline{0.1916} & \textbf{0.5371} & \textbf{0.9143} & \underline{0.0223} \\
$\uparrow$ Echo cancellation & Speech & SDR & \underline{2.9642} & 0.8051 & \textbf{9.7490} & -1.6403 & \textbf{0.1448} & \underline{-0.7459} & nan & \textbf{8.6398} \\
$\uparrow$ Filtering Speech & Speech-multi peak-NS & SDR & \textbf{11.7282} & \underline{11.3844} & 10.0130 & 9.7817 & \textbf{11.1821} & \underline{10.1974} & nan & \textbf{15.4746} \\
$\uparrow$ Filtering Speech & Speech-single peak-S & SDR & 5.4157 & 3.9530 & \underline{7.6142} & \textbf{16.1273} & \underline{4.2970} & \textbf{5.3683} & \underline{0.6694} & \textbf{8.1903} \\
$\uparrow$ Filtering Speech & Speech-multi peak-S & SDR & 5.4320 & 5.2824 & \underline{5.4885} & \textbf{6.4630} & \underline{3.4378} & \textbf{4.6160} & \textbf{13.5152} & \underline{5.6517} \\
$\uparrow$ Filtering Speech & Speech-single peak-NS & SDR & 3.1390 & 4.3511 & \underline{4.4404} & \textbf{6.4976} & \underline{0.7624} & \textbf{1.3685} & nan & \textbf{-0.4859} \\
\midrule
$\downarrow$ Imputation & ECG & MSE & \textbf{0.0201} & \textbf{0.0201} & \textbf{0.0201} & \underline{0.0230} & \textbf{0.0201} & \underline{0.0245} & nan & \textbf{0.0246} \\
$\downarrow$ Imputation & PPG & MSE & \textbf{0.3182} & 1.5787 & \textbf{0.3182} & \underline{1.2262} & \underline{0.3051} & \textbf{0.2752} & nan & \textbf{0.3870} \\
$\downarrow$ Filtering ECG & ECG-60Hz & MSE & 0.0040 & \underline{0.0039} & \underline{0.0039} & \textbf{0.0000} & \textbf{0.0028} & \underline{0.0073} & \textbf{0.0057} & \underline{0.0070} \\
$\downarrow$ Filtering ECG & ECG-50Hz & MSE & \textbf{0.0000} & \textbf{0.0000} & \textbf{0.0000} & \textbf{0.0000} & \textbf{0.0021} & \underline{0.0091} & \underline{0.0247} & \textbf{0.0198} \\
$\downarrow$ Filtering ECG & ECG-motion & MSE & \underline{0.0001} & 0.0046 & \textbf{0.0001} & 0.0001 & \textbf{0.0052} & \underline{0.0138} & nan & \textbf{0.0065} \\
$\downarrow$ Filtering ECG & ECG-gaussian & MSE & \underline{0.0023} & 0.0026 & \textbf{0.0007} & 0.0027 & \underline{0.0120} & \textbf{0.0017} & \underline{0.0183} & \textbf{0.0083} \\
$\downarrow$ Extrapolation & PPG & MSE & \underline{0.5148} & \textbf{0.4217} & 0.6087 & 2.5117 & \textbf{0.4242} & \underline{0.5594} & \textbf{0.0918} & \underline{809.7639} \\
$\downarrow$ Extrapolation & ECG & MSE & 0.0222 & \textbf{0.0147} & 0.0222 & \underline{0.0193} & \textbf{0.0144} & \underline{0.0219} & \underline{0.4613} & \textbf{0.1756} \\
$\downarrow$ ECG heartrate  & ECG & MSE & 1546.0656 & \underline{667.8772} & \underline{667.8772} & \textbf{36.2437} & \textbf{3.87e+06} & \underline{1.30e+07} & nan & \textbf{1.49e+08} \\
$\downarrow$ Delay detection & Gait & MSE & \textbf{0.0006} & 0.0323 & 0.0091 & \underline{0.0020} & \underline{2.3273} & \textbf{0.0017} & \textbf{16.2394} & \underline{87345.2444} \\
$\downarrow$ Period detection & Gait & MSE & 0.3661 & 0.4279 & \underline{0.3589} & \textbf{0.2139} & \textbf{0.6990} & \underline{0.8127} & \underline{1.06e+15} & \textbf{16800.2540} \\
$\downarrow$ Resampling  & Synthesis & MSE & \textbf{0.0003} & 0.0535 & 0.1242 & \underline{0.0366} & \textbf{0.0642} & \underline{0.0889} & \textbf{0.7537} & \underline{0.8662} \\
\midrule
Win rate (\%) & - & - & 24 & 24 & \underline{28} & \textbf{48} & \textbf{56} & \underline{44} & \underline{24} & \textbf{76}  \\
Failure rate (\%) & - & -& \underline{12.85} & 14.23 & 13.54 & \textbf{5.75} & \underline{23.61} & \textbf{17.86} & \underline{84.40} & \textbf{37.56}  \\
\bottomrule
\end{tabular}
}
\caption{Comparisons among all the methods, including base, CoT, ReAct, self-verification on GPT-4o, and fine-tuning on Llama 70B and 3B. The best is made \textbf{bold}, and the second best is \underline{underscored} when compared within the same LLM. Note that higher F1 and SDR scores indicate better performance (marked with $\uparrow$).}
\label{table:prompt-eng}
\vspace{-15pt}
\end{table*}






\subsection{Baselines}
We consider supervised fine-tuning and four prompt engineering methods: Base, Chain-of-Thought (CoT), ReACT, and self-verification.


\noindent{\bf Base.} This represents the simplest form of interaction with the LLM, involving straightforward queries without additional prompting techniques. \looseness=-1

\noindent{\bf CoT\cite{wei2022chain}.} Chain of Thought (CoT) prompting involves guiding the model through a series of logical steps to reach a solution. This method encourage the model to break down complex tasks into smaller and manageable steps.

\noindent{\bf ReAct\cite{yao2022react}.} ReAct is a prompt engineering strategy where the model is encouraged to reflect on its previous outputs and actions before proceeding with the next step. This method aims to simulate an iterative problem-solving approach.




\noindent{\bf Self-verification\cite{shinn2024reflexion}} asks LLMs to propose tentative solutions, obtain initial feedback, and then refine their strategies based on task feedback. Building on \cite{shinn2024reflexion}, we employ another LLM as an internal sanity checker to provide autonomous feedback, and we set the maximum self-verification rounds to be 5. Implementation details are available in Appendix \ref{Sec:llmdsp}.


\noindent{\bf Fine-tuning.} We finetune open-source LLMs on a self-collected instruction-tuning dataset. The datasets comprise 11266 query-response pairs generated based on open-source codebases. We selected 16 high-quality open-source codebases and tutorials from \textit{Python} and MATLAB libraries. Then, we employed GPT-4o-mini to generate coding challenges from the sources, following the procedures outlined in \cite{dubey2024llama, wei2024magicoder}. We then applied LoRA-based fine-tuning techniques \cite{hu2021lora} to both the Llama-3.1-70B model and the Llama-3.2-3B model\cite{dubey2024llama}. Additional details about the dataset preparation and fine-tuning procedures can be found in Appendix \ref{appendix:finetune}.

\subsection{Results}

Table \ref{table:prompt-eng} shows that Self-verification emerges as the most effective approach, achieving the highest win rate (48\%) and lowest failure rate (5.75\%) among all methods. This suggests that iterative self-correction is particularly beneficial for sensor processing tasks. For example, Figure \ref{fig_filer_example} illustrates an example of invalid assumptions made by a model. While the model effectively uses \emph{scipy}, it neglects the critical selection of the stop-band frequency. Self-verification, mimicking an engineer’s trial-and-error approach, helps correct such errors. However, we observe that the effectiveness of self-verification also relies on the difficulties of defining a valid verification criterion. The models could make mistakes in self-verification, leading to diminished performance, if a valid criterion is hard to define. Appendix \ref{Sec:impact-self-verif} provides task-wise correlations between self-verification and the performance of models, highlighting the effects of the approach.




\begin{figure}[t] 
\begin{tcolorbox}[
    colback=blue!10!white,
    colframe=blue!50!black,
    label=fig_filer_example, 
    fonttitle=\bfseries,
]
\begin{lstlisting}
import numpy as np
from scipy.signal import butter, lfilter, freqz
    ...
    # Determine the siren frequency range by analyzing the input signal
    # For simplicity, assuming the siren is within the range 1600 - 2400 Hz
    lowcut = 1600
    highcut = 2400
    filtered_data = bandstop_filter(input_data, lowcut, highcut, sampling_rate)
    return filtered_data
\end{lstlisting}
\end{tcolorbox}
\vspace{-10pt}
\caption{Model output example. The model makes an invalid assumption on stop-band frequency.}\label{fig_filer_example}
\vspace{-20pt}
\end{figure}

Additionally, Fine-tuning results vary with model size. For the larger Llama-70B model, fine-tuning shows limited improvements, possibly due to the model's already extensive training on coding tasks. In contrast, fine-tuning significantly improves the smaller Llama-3B model's performance, reducing failure rates and increasing win rates. A possible explanation for this discrepancy is that the Llama-70B model has already been extensively trained on a rich coding corpus. As a result, injecting DSP knowledge through fine-tuning with the sensor processing corpus offers limited performance improvements. Conversely, the smaller model benefits more from this fine-tuning process.


Based on the study above, we summarize our response to Q2: Self-verification outperforms other prompting strategies on the majority of tasks. Mimicking human experts through iterations of problem-solving can enhance LLM performance, while the effectiveness of fine-tuning is limited. However, there is still a considerable performance gap between LLMs and human experts. Even with the best prompt engineering approach, the \name is outperformed by Expert 1 in 72\% of tasks and by Expert 2 in 68\% of tasks.\looseness=-1

\section{Conclusion and Discussion}

In this work, we systematically investigated the potential of LLMs to function as sensor-processing \emph{copilot}. Our study begins with introducing a comprehensive benchmark to systematically analyze the LLMs' signal processing abilities for sensing tasks.\looseness=-1

Our findings reveal that while LLMs demonstrate competence in simpler tasks, there is a significant performance gap when it comes to more challenging tasks. Specifically, our evaluation indicates that LLMs underperform by over 50\% on complex tasks compared to human experts. 

Despite these challenges, it is worth noting that the complex tasks in \bchname are also challenging for human experts, who often spend hours solving them effectively. This highlights the inherent difficulty of the tasks rather than the inadequacy of the models. Our work represents a step towards making signal processing more accessible, which would otherwise remain daunting for non-experts. 


Finally, the inferior performance of LLMs in compositional and parameterized tasks suggests that LLMs primarily perform knowledge retrieval from their training data. This highlights that the key to enabling LLMs as effective sensor processing \emph{copilots} lies not in fine-tuning or more training data but in redesigning their core architecture to improve reasoning and planning abilities. The self-verification and test-time scaling designs \cite{snell2024scaling} seem to be promising directions.

In conclusion, our work aspires to set a clear objective for this rapidly evolving field and drive progress toward reliable sensor processing solutions powered by LLMs.
\section{Limitations}
This work focuses on single-channel temporal sensor signals, such as audio, ECG, PPG, motion, and pressure signals. Future work will explore spatial-temporal and multimodal sensor signals and other common sensor processing problems.

Besides, this work has not touched other forms of interactions to incorporate temporal signals to LLMs, including quantization-based methods that convert numerical data into discrete representations \cite{rabanser2005effectiveness}, feeding temporal signals to an encoder to align the signal with texts \cite{jin2023time}, or interpreting temporal signals with visuals, e.g., line graphs or spectrogram \cite{wimmer2023leveraging}. These approaches are promising in enhancing LLMs' understanding of raw sensory signals, and we defer their exploration to future works.

Another limitation of this study is the inclusion of only three engineering experts as the human baseline. This constraint arises from the scarcity of individuals with specialized engineering expertise and the difficulty in recruiting such experts through conventional crowdsourcing platforms. As a result, the limited number of expert participants may affect the generalizability of the findings.
\section{Acknowledgment}


The research reported in this paper was sponsored in part by: the Air Force Office of Scientific Research under Cooperative Agreement \#FA95502210193; the DEVCOM Army Research Laboratory under Cooperative Agreement \#W911NF-17-2-0196; the National Science Foundation under Award \#CNS-2325956, and, the National Institutes of Health under Award \#1P41EB028242.  Any findings in this material are those of the author(s) and do not reflect the views of any of the above funding agencies. The U.S. Government is authorized to reproduce and distribute reprints for Government purposes notwithstanding any copyright notation here on.

The authors would like to thank OpenAI for their support through the OpenAI Researcher Access Program.

\bibliography{reference}
\bibliographystyle{acl_natbib}

\appendix
\section{Appendix}
\label{sec:appendix}

\subsection{Performance under Different Settings}\label{appendix:settings}

\begin{figure*}
    \centering
 \setlength{\abovecaptionskip}{0.cm}
    \setlength{\belowcaptionskip}{0.cm}
    \begin{subfigure}[b]{0.32\textwidth}
        \includegraphics[width=\textwidth]{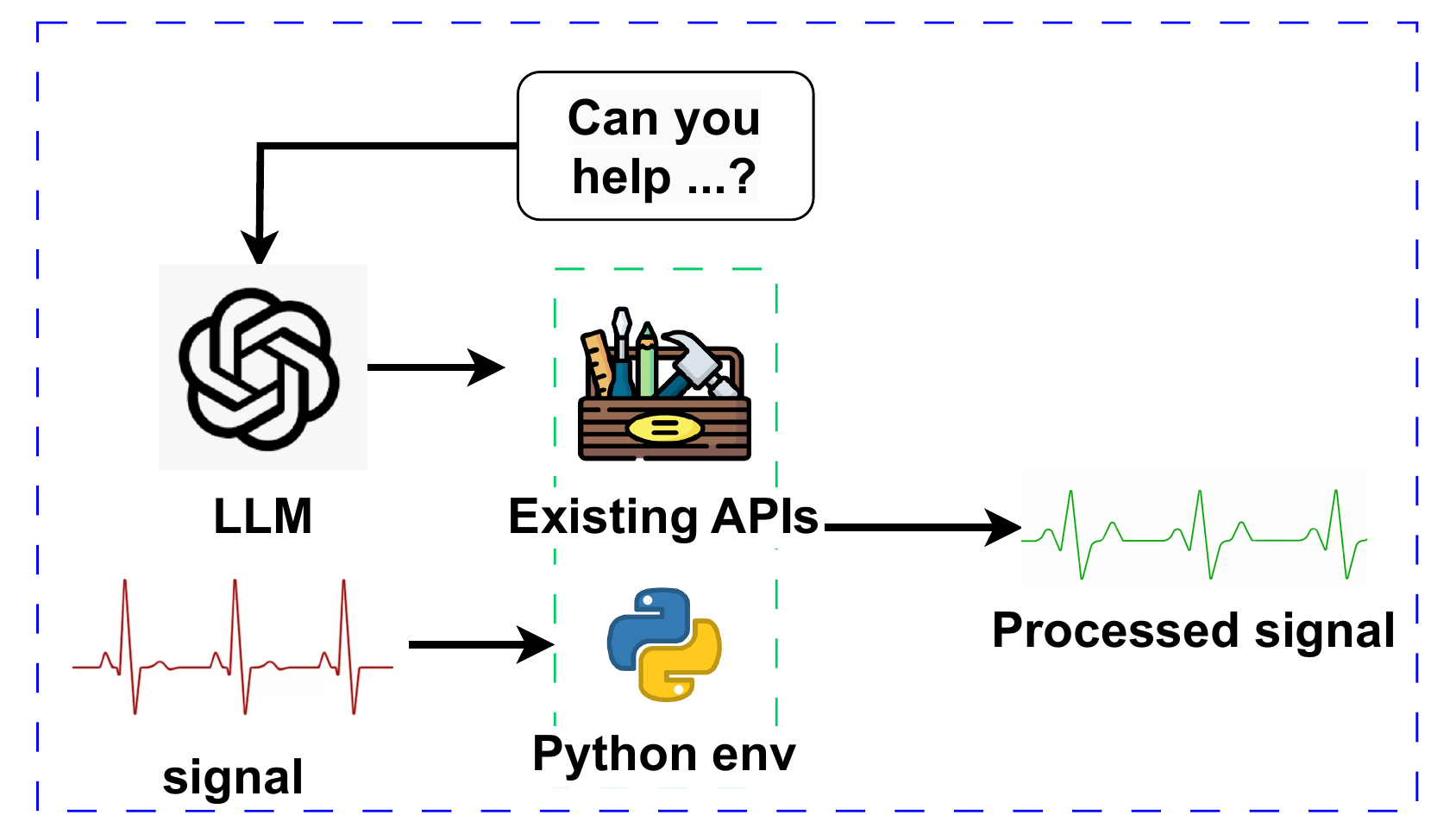 }
        \caption{Coding w/ APIs}
    \end{subfigure}
    \begin{subfigure}[b]{0.32\textwidth}
        \includegraphics[width=\textwidth]{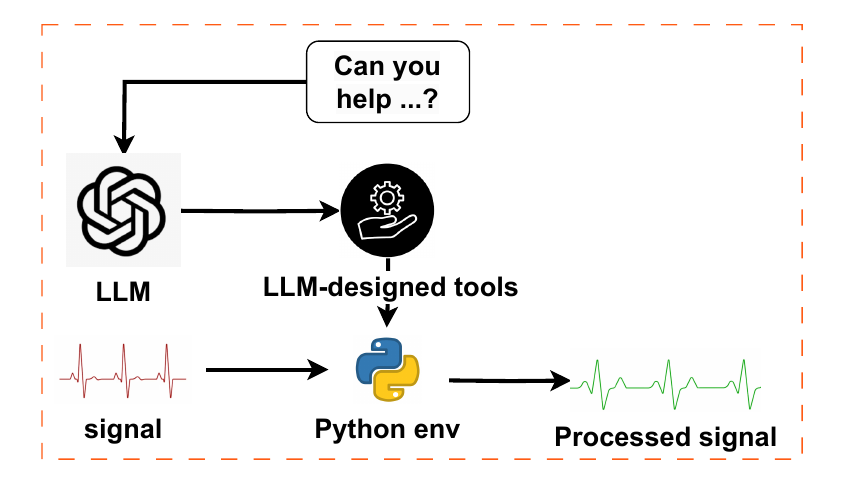 }
        \caption{Coding w/o. APIs}
    \end{subfigure}
    \begin{subfigure}[b]{0.32\textwidth}
        \includegraphics[width=\textwidth]{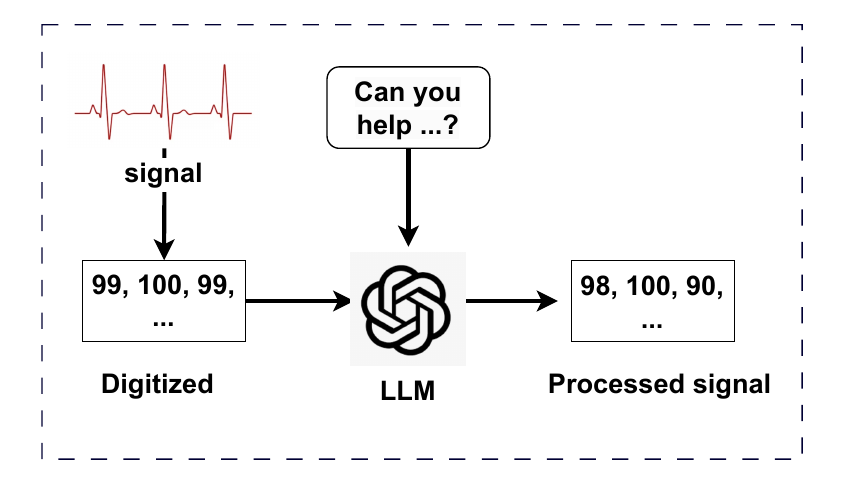 }
        \caption{Text-based}
    \end{subfigure}

    \caption{Settings of using LLMs in DSP: coding with APIs, coding without APIs, and text-based interactions.} 
    \label{fig:setting_LLMs}
    
\end{figure*}

To explore using LLMs for sensor processing, we designed three distinct interaction settings that mimic expert engineers' ways of processing sensors. Expert engineers typically approach a problem by either calling existing APIs, coding without APIs (creating their own tools), or solving the problem directly and outputting results. These strategies informed our experimental settings as shown in Fig. \ref{fig:setting_LLMs}: (1) coding with APIs, where LLMs leverage pre-existing libraries and functions to address DSP tasks (the defined list of available APIs includes \emph{numpy}, \emph{scipy}, \emph{pandas}, \emph{pmdarima}, \emph{statsmodels}, and \emph{ruptures}), (2) coding without APIs, where LLMs generate custom code solutions independently of external resources, and (3) text-based interaction, where LLMs tackle DSP challenges through conversation with digitized signals. By structuring our investigation around these scenarios, we benchmark the performance of the four different LLMs, GPT-4o \cite{gpt_4o}, GPT-4 \cite{achiam2023gpt}, GPT-3.5-turbo \cite{gpt_3_5_turbo}, and Llama-3-70b \cite{meta_llama_3}. The prompts of the settings are in Appendix \ref{section:prompt}.

We show the model performance on three types of settings: coding with APIs (denoted as A), coding without APIs (denoted as NA), and text-based (denoted as T). Table \ref{table:setings} shows the performance for different combinations of models and interacting approaches.




From Table \ref{table:setings}, we can observe that coding with APIs outperforms other ways of interactions consistently. Among the models, Llama-3-70b has the highest win rate (68.0\%) in the coding with APIs setting, followed by GPT-4o (60.0\%). 

From Table \ref{table:setings_taskwise}, we can observe that coding with APIs almost achieves the highest win rates for the models in all runs. Occasionally, outlier detection is performed better using coding w/o APIs. The reason is that comparing with simple standard deviation is sufficient in some outlier detection tasks, and thus, it is better to use a simple implementation. 

Lastly, text-based interaction is mostly worse than coding with APIs and coding without APIs. The only exception is the extrapolation. This demonstrates the capabilities of the text model doing time series reconstruction tasks shown in \cite{mirchandani2023large}. 

Based on the results, we conclude that the current LLMs perform the best by coding with APIs for sensor signal processing. Coding with APIs significantly outperforms others consistently and shows the same trend for all models we tested.


\begin{table}
\centering
\resizebox{0.35\textwidth}{!}{%

\begin{tabular}{lccc}
\toprule
Task & A  & NA & T \\
\midrule
Change point detection & \textbf{56.3} & 12.5 & \underline{31.2} \\
Delay detection & \textbf{75.0} & 0.0 & \underline{25.0} \\
ECG heart rate & \textbf{50.0} & 0.0 & 50.0 \\
Echo cancellation & \textbf{75.0} & \underline{25.0} & 0.0 \\
Extrapolation & \underline{37.5} & 12.5 & \textbf{50.0} \\
ECG Filtering & \textbf{81.2} & \underline{18.8} & 0.0 \\
Speech Filtering & \textbf{75.0} & \underline{25.0} & 0.0 \\
Imputation & \textbf{50.0} & \underline{37.5} & 12.5 \\
Outlier detection & \underline{31.2} & \textbf{50.0} & 18.8 \\
Period detection & \textbf{50.0} & 25.0 & \underline{25.0} \\
Resampling & \textbf{100.0} & 0.0 & \underline{0.0} \\
\bottomrule
\end{tabular}

}
\caption{Win rate for coding with APIs (A), coding without APIs (NA), and text-based (T). Numbers are averaged among all the models. The best is made \textbf{bold}, and the second best is \underline{underscored}.}
\label{table:setings_taskwise}\vspace{-1em}
\end{table}

\section{Self-supervised Finetuning Methodology}\label{appendix:finetune}

To fine-tune the models, we adopted a structured pipeline following \cite{dubey2024llama, wei2024magicoder}.
\begin{enumerate}[leftmargin=*]
    \item \textbf{Dataset preparation} was conducted by crawling established signal processing tutorials, such as MATLAB signal processing resources, Scipy documentation, and Scipy example repositories in Table \ref{tab:opensource_codebases_licenses}. This resulted in a corpus comprising text-code example pairs, Jupyter Notebook files, and open-source digital signal processing (DSP) implementations.
    \item \textbf{Problem description generation}. We used the GPT-4o-mini model to generate programming problems inspired by the extracted code snippets. Each problem was designed to align with the corresponding example, ensuring relevance and context.
    \item \textbf{Solution generation}. Next, GPT-4o-mini was prompted to provide solutions to the generated problems in Python. The model is instructed to only leverage libraries permitted in this paper.
    \item \textbf{Correctness analysis and refinement}. We perform solution checking for syntax errors and execute the code to identify runtime issues, such as uninitialized variables or missing imports. Finally, an iterative self-correction mechanism was employed to refine solutions. If a solution failed any correctness checks, the model was re-prompted to revise it, with iterations continuing until an executable code snippet was achieved. 
\end{enumerate}

As for fine-tuning, we use the LoRA implementation by Unsloth \cite{unsloth}. We set the learning rate to $2\mathrm{e}{-4}$, warmup step to $5$, batch size to $8$, weight decay to $0.01$, LoRA alpha to 16, and rank to 16. Finetuning is performed on NVIDIA H100 GPUs with 96 GB VRAM for 2 epochs.


\begin{table}[ht]
\centering
\resizebox{\linewidth}{!}{%
\begin{tabular}{l|l}
\toprule
\textbf{Resource name} & \textbf{License Info.} \\ 
\midrule
\href{https://docs.scipy.org/doc/scipy/tutorial/interpolate/extrapolation_examples.html}{Scipy Documentation Examples} & BSD 3-Clause \\
\hline
\href{https://www.mathworks.com/help/signal/examples.html}{MATLAB Signal Processing Toolbox} & MathWorks \\
\hline
\href{https://github.com/AllenDowney/ThinkDSP}{ThinkDSP Textbook Examples} &  MIT \\
\hline
\href{https://github.com/scipy/scipy-cookbook}{Scipy cookbooks repository} & BSD 3-Clause \\
\hline
\href{https://scipy-cookbook.readthedocs.io/items/AccumarrayLike.html}{Scipy cookbooks} & BSD 3-Clause \\
\hline
\href{https://pysdr.org/content/sampling.html}{PySDR Book} & CC BY-NC 4.0 \\
\hline
\href{https://github.com/jinglescode/python-signal-processing}{Python Signal Processing} & MIT \\
\hline
\href{https://github.com/MIT-LCP/wfdb-python}{WFDB Toolbox for Python} & MIT \\
\hline
\href{https://github.com/mne-tools/mne-python}{MNE-Python for EEG Analysis} & BSD 3-Clause \\
\hline
\href{https://github.com/neuropsychology/NeuroKit}{NeuroKit Physiological Analysis} & MIT \\
\hline
\href{https://github.com/Gabrock94/Pysiology.git}{Pysiology Toolbox} & MIT \\
\hline
\href{https://github.com/PGomes92/pyhrv.git}{pyHRV: Heart Rate Variability Analysis} & GPLv3 \\
\hline
\href{https://gitlab.com/a.bizzego/pyphysio.git}{PyPhysio for Physiological Signal Analysis} & MIT \\
\hline
\href{https://github.com/paulvangentcom/heartrate_analysis_python.git}{Heart Rate Analysis in Python} & MIT \\
\hline
\href{https://github.com/scientisst/BioSPPy.git}{BioSPPy: Biosignal Processing} & BSD 3-Clause \\
\hline
\href{https://github.com/PyWavelets/pywt.git}{PyWavelets: Wavelet Transforms} & MIT \\
\bottomrule
\end{tabular}
}
\caption{Open-source codebases used to generate fine-tuning corpus with license information.}
\label{tab:opensource_codebases_licenses}
\end{table}
\section{Prompts}\label{section:prompt}

\begin{figure*}[t] 
\begin{tcolorbox}[
    colback=blue!10!white,
    colframe=blue!50!black,
    title=Prompt of text-based setting, 
    label=promt_text, 
    fonttitle=\bfseries
]
You are an AI model good at understanding, manipulating, and modifying sensory data without resorting to any tools. 

You are capable of handling users' requests on a series of signal-processing tasks, such as prediction, imputation, filtering, and detection.

\end{tcolorbox}
\caption{Prompt of text-based setting}
\end{figure*}

\begin{figure*}[t] 
\begin{tcolorbox}[
    colback=blue!10!white,
    colframe=blue!50!black,
    title=Prompt of coding without APIs setting, 
    label=promt_no_api, 
    fonttitle=\bfseries
]
You are an expert in signal processing. Your role is to analyze and process various types of signals (such as audio, electromagnetic, or physiological signals) using your Python coding. You are expected to process signal directly without user interference.

Instructions:

1. Use Python codinng for signal processing tasks. Implement your functions inside ```Python ``` code block. Do not write code outside the functions. The function prototypes are as follows:

You just need to implement the function the solver (mandatory):
\\

\begin{lstlisting}
 ```Python 
def solver(input_data, sampling_rate=None):
    # HERE is where you put your solution
    # Args:
    #   input_data: The data type is numpy.ndarray. This is the data provided by the user to perform DSP. 
    #   sampling_rate: The sampling rate of the data. sampling_rate is mandatory for speech, ecg, ppg, and gait data. It could be optional for others.
    # Output:
    #   return: return the processed data in numpy.ndarray
 ```
 \end{lstlisting}

Please note that variables input\_data and sampling\_rate are provided through the function solver. Do not simulate them or write code outside the designated function.

2. Iterative problem solving: first state the key ideas to answer user's query and solve the problem iteratively (do not over-divide the steps).

3. [IMPORTANT] Specific Interactive Format: Users will put their queries into the format \\QUERY[text]. For example, \\QUERY[Can you denoise my ECG signal that's corrupted by powerline noise?]. When you finished, state the keyword [SUCCESS], and the iteration will stop. Output [SUCCESS] in the chat directly. 

4. [IMPORTANT] Use your own implementation: You should implement the functions w/o relying on APIs other than numpy. Do not use scipy.

For instance, if you want to perform spectral filter, you should come up with your own implementation.

Now I am going to provide the query to you, and you need to start answering the query using Python. Say "I am ready" if you understand the problem.
\end{tcolorbox}
\caption{Prompt of coding without APIs setting}
\end{figure*}

\begin{figure*}[t] 
\begin{tcolorbox}[
    colback=blue!10!white,
    colframe=blue!50!black,
    title=Prompt of coding with APIs setting, 
    label=promt_api, 
    fonttitle=\bfseries
]
You are an expert in signal processing. Your role is to analyze and process various types of signals (such as audio, electromagnetic, or physiological signals) using your Python coding. You are expected to process signal directly without user interference.

Instructions:

1. Python Coding: Use Python codinng for signal processing tasks. Implement your functions inside ```Python ``` code block. Do not write code outside the functions. The function prototypes are as follows:

You just need to implement the function the solver (mandatory):
\\

\begin{lstlisting}
 ```Python 
def solver(input_data, sampling_rate=None):
    # HERE is where you put your solution
    # Args:
    #   input_data: The data type is numpy.ndarray. This is the data provided by the user to perform DSP. 
    #   sampling_rate: The sampling rate of the data. sampling_rate is mandatory for speech, ecg, ppg, and gait data. It could be optional for others.
    # Output:
    #   return: return the processed data in numpy.ndarray
 ```
\end{lstlisting}

Please note that variables input\_data and sampling\_rate are provided through the function API. Do not simulate them or write code outside the designated function.

2. Iterative problem solving: first state the key ideas to answer user's query and solve the problem iteratively (do not over-divide the steps).

3. [IMPORTANT] Specific Interactive Format: State all your output directly. DO NOT put it inside code or with ```. Users will put their queries into the format \\QUERY[text]. For example, \\QUERY[Can you denoise my ECG signal that's corrupted by powerline noise?]. When you finished, state the keyword [SUCCESS], and the iteration will stop. Output [SUCCESS] in the chat directly. 

4. [IMPORTANT] Remember, you are a text-based model. You shouldn't inspect visual or listen to audios directly (e.g., write code to visualize them). To understand a signal, you need to interact through text or design methods to learn about the properties.

End Goal: Your ultimate goal is to provide independent, accurate, and accessible signal-processing assistance, achieving their objectives efficiently and effectively.

You can make use of the following libraries (Do NOT attempt to install new libraries):

(1) numpy: Numpy provides mathematical operations on signals, such as array manipulation Fourier transforms, statistical analysis.
(2) scipy: Scipy is generally useful for filter design, signal transformation, and signal analysis. You can use the libraries from ```Python scipy.signal``` for filter design. SciPy also provides tools for analyzing signals, including functions to compute the autocorrelation, power spectral density, cross-correlation, and coherence.
(3) pandas: Pandas is useful for time series data manipulation and analysis. For example, you can use ```Python pandas.Series``` to compute rolling mean or standard deviation.
(4) pmdarima: Pmdarima is primarily used for time series forecasting, while it can also be extended to Seasonal Decomposition, Anomaly Detection, or Data Smoothing.
(5) statsmodels: Statsmodels is a Python library designed for statistical modeling, estimation, and testing, while it can also be extended spectral analysis and filtering.
(6) ruptures: Ruptures is to identify points in time where the statistical properties of a signal or time series change.

\end{tcolorbox}
\caption{Prompt of coding with APIs setting}
\end{figure*}

\begin{figure*}[t] 
\begin{tcolorbox}[
    colback=blue!10!white,
    colframe=blue!50!black,
    title=Prompt of Reflector, 
    label=promt_text, 
    fonttitle=\bfseries
]
You are an advanced signal processing agent that can perform reflection on a signal processing plan. You are tasked with another text-based signal-processing AI that handles signal-processing queries by planning and coding. Your job is to reflect on the previous plan. Do not intend to use tools that are not specified. You will be given the previous signal processing trial as context, the user's query, and The AI's previous performance. \
First, diagnose if the previous execution is a successful workaround to the query. If yes, output [SUCCESS], and the iteration will stop.
If not, output [FAILED] and start proposing a possible reason for failure and devise a new, concise, high-level plan. 

[important] Reflect instruction:
(1) Be specific in your feedback. Give detailed examples of where the AI makes mistakes.
(2) Be careful if the model selected parameters or performed steps carelessly. Provide a revised plan to rectify this.
(3) Check if the model makes unrealistic or incorrect assumptions.
(4) Do not suggest libraries that the AI agent should not use.
(5) Remember, Both you and the other AI models are text-based. Both shouldn't inspect visuals or listen to audio directly. Check if the other model tries to plot or hear signals using Python directly. If so, point that out and ask the model to use external functions to understand the signals.
(6) Finally, an external expert will give a performance evaluation of the AI agent's output. Combine evaluation and the AI agent's code to determine whether it is [SUCCESS] or [FAILED].

[important] Reflection format:
[SUCCESS]\/[FAILED]: First, state [SUCCESS] or [FAILED]
[Summary]: Second, give a brief summary of the outline of the previous attempt. 
[Analysis]: Then, state one major reason for failure in the last attempt. Specify which part of the previous code was wrong. 
[Revised Plan]: Finally, state what to do to improve. Do not write Python code directly. Do not overthink. Make it succinct and accurate. 

Here is the previous trial information:
[Relevant CONTEXT STARTS]: \{context\} [CONTEXT ENDS.]

[Question]: \{question\}

[Previous Performance]: \{performance\}

[Performance hist]: \{performance\_hist\}

Now, start your reflection by judging [SUCCESS]/[FAILED] from the previous attempt and then begin your reflection:
\end{tcolorbox}
\caption{Prompt of Reflector.}
\end{figure*}

\begin{figure*}[t] 
\begin{tcolorbox}[
    colback=blue!10!white,
    colframe=blue!50!black,
    title=Prompt of Verifier, 
    label=promt_text, 
    fonttitle=\bfseries
]
You are a verifier that can perform evaluation on a signal processing plan. \
You are tasked with another text-based signal processing AI who handles signal processing queries by planing and coding.\
Your job is to perform a sanity check on the results using Python.\
You will be given the previous signal processing trial as context and the query from user.

You have access to the following libraries:

...

[important] Evaluation protocal:
- Do it in four steps in the following format.
- [INSPECTION]: First, inspect the output\_data by writing the following function. Check its validity. If not valid, directly output False in the function. And the iteration stops. 

    1) The function prototype is as follows:
    2) Please note that variables input\_data, output\_data, and sampling\_rate is accessible through the function interface. Do NOT simulate them on your own.
\\
\begin{lstlisting}
```Python
def inspection(input_data, output_data, sampling_rate=None):
    # Inspect the output_data and output True/False. 
    # 1) Check if the output_data has the valid range, is empty, or contains missing values. 
    # 2) Do NOT check the data type - using the isinstance or np.isscalar function is not reliable.
    # Args:
    #   input_data: The data type is numpy.ndarray. This is the data provided by the user to perform DSP. 
    #   output_data: The data type is numpy.ndarray. The variable is provided through the function interface for you. This is the data processed by the other AI agent. 
    #   sampling_rate: The sampling rate of the data. sampling_rate is mandatory for speech, ecg, ppg, and gait data. It could be optional for others.
    # Output: boolean variable - True or False. If the result does not pass your test, output False. Else, output True.
```
\end{lstlisting}
...
    
- [ANALYSIS]: Based on your goal, only implement the challenger function to verify if it is true. Use data provided by the user and the output data produced by the AI agent through challenger API.
    1) Remember, you are a language model. Do not directly plot signals and inspect them or hear audios.
    2) Do not reproduce the solver function. Instead, you should check if the output\_data satisfy some properties.
\\
\begin{lstlisting}
```Python 
def challenger(input_data, output_data, sampling_rate=None):
    # HERE is where you put your sanity check code. 
    # Args:
    #   input_data: The data type is numpy.ndarray. The variable is provided through the function interface for you. This is the data provided by the user to perform DSP. 
    #   output_data: The data type is numpy.ndarray. The variable is provided through the function interface for you. This is the data processed by the other AI agent. 
    #   sampling_rate: The sampling rate of the data. sampling_rate is mandatory for speech, ecg, ppg, and gait data. It could be optional for others.
    # Return: boolean variable - True or False. If your the result does not pass your test, output False. Else, output True.
 ```
 \end{lstlisting}
 
    ...
    
Here is the previous trial information:
[Relevant CONTEXT STARTS]: {context} [CONTEXT ENDS.]
[Question]: {question}
[Memory]: {memory}
[output\_data]: {vis\_result}

Now, start your evaluation step by step:
\end{tcolorbox}
\caption{Prompt of Verifier.}
\end{figure*}




\end{document}